\documentclass[11pt,a4paper]{article}
\usepackage[hyperref]{acl2020}
\usepackage{times}
\usepackage{latexsym}

\usepackage{hyperref}
\usepackage{url}
\usepackage{comment}
\usepackage{graphicx}
\usepackage{multirow}
\usepackage{multicol}
\usepackage{array}
\usepackage{float}
\usepackage{colortbl}
\definecolor{mygray}{gray}{.9}
\usepackage{microtype}

\aclfinalcopy 

\title{Data Annealing for Informal Language Understanding Tasks}

\author{Jing Gu \\
  University of California, Davis \\
  \texttt{jkgu@ucdavis.edu} \\\And
  Zhou Yu \\
  University of California, Davis \\
  \texttt{joyu@ucdavis.edu} \\}

\date{}

\begin{document}
\maketitle

\begin{abstract}
\label{abstract}
There is a huge performance gap between formal and informal language understanding tasks. The recent pre-trained models that improved the performance of formal language understanding tasks did not achieve a comparable result on informal language.
We propose a data annealing transfer learning procedure to bridge the performance gap on informal natural language understanding tasks. It successfully utilizes a pre-trained model such as BERT in informal language.
In our data annealing procedure, the training set contains mainly formal text data at first; then, the proportion of the informal text data is gradually increased during the training process. Our data annealing procedure is model-independent and can be applied to various tasks.
We validate its effectiveness on exhaustive experiments. 
When BERT is implemented with our learning procedure, it outperforms all the state-of-the-art models on the three common informal language tasks.
\end{abstract}

\section{Introduction and Related Work}
\label{sec:intro}
Because of the noisy nature of the informal language and the shortage of labelled data, the progress on informal language is not as promising as on formal language.
Many tasks on formal data obtain a high performance due to deep neural models \citep{lstm_transfer_ner, elmo, bert-paper}. However, usually, these state-of-the-art models' excellent performance can not directly transfer to informal data. 
For example, when a BERT model is fine-tuned on informal data, its performance is less encouraging than on formal data.
This is caused by the domain discrepancy between the pre-training corpus used by BERT and the target data.
%



To solve the issues mentioned above, we propose a model-agnostic data annealing procedure. The core idea of data annealing is to let the model have more freedom to explore its update direction at the beginning of the training process.
More specifically, we set informal data as target data, and we set formal data as source data. 
The training data first contains mainly source data, when data annealing procedure takes the advantages of a good parameter initialization from the clean nature of formal data.
Then the proportion of source data keeps decreasing while the proportion of target data keeps increasing. 
Recent works have validated the effect on the changing of data type during training period. Curriculum learning suggests a proper order of training data improves the performance and speed up training process in single dataset setting \cite{curriculum_learning}. Researchers have also proven the effect of data selection in domain adaptation \citep{learning_to_select_data, data_selection, data_select_translation}. 

The philosophy behind data annealing is shared with other commonly used annealing techniques.
One popular usages of annealing is setting learning rate of the neural model. A gradually decayed learning rate gives the model more freedom of exploration at the beginning and leads to better model performance \citep{adadelta_use_annealing_learning_rate, ncrf++, bert-paper}.
Another popular implementation of annealing is simulated annealing \citep{simulated_annealing}. It reduces the probability of a model converging to a bad local optimal by introducing random noise in the training process.
Data annealing has similar functionality with simulated annealing. Data annealing replaces random noise with source data. By doing this, the model is not only able to explore more space at the beginning of the training process, but also the model is guided by the knowledge learned from the source domain. \par

Current state-of-the-art models on informal language tasks are usually designed specifically for certain task and cannot generalize to different tasks \cite{twitter_hate_speech, POS_previous_best}.  
Data annealing is model-independent and therefore could be employed on different informal language tasks.
We validate our learning procedure with two popular neural network models in NLP, LSTM and BERT, on three popular natural language understanding tasks, i.e., named entity recognition (NER), part-of-speech (POS) tagging and chunking on twitter.
When BERT is fine-tuned with data annealing procedure, it outperforms all three state-of-the-art models with the same structure. By doing this, we also set the new state-of-the-art result for the three informal language understanding tasks.
Experiments also validate the effectiveness of our data annealing procedure when there are limited training resources in target data.

\section{Data Annealing}
\label{methods_section}
A pre-trained model like BERT is suggested to avoid over-training when implemented on downstream task \citep{pretrain_or_not, bert_forgetting}. 
It is not ideal to feed too much source data, as it not only prolongs the training time but also confuses the model.  Therefore, we propose data annealing, a transfer learning procedure that adjusts the ratio of the formal source data and the informal target data from large to small in training process to solve the overfitting and the noisy initialization problems. 

At the beginning of the training process, most of the training samples are source data. Therefore the model obtains a good initialization from the abundant clean source data.
We then gradually increase the proportion of the target data and reduce the proportion of the source data. Therefore, the model explores a larger parameter space. Besides, the labelled source dataset works as an auxiliary task.
At the end of the training process, we let most of the training data to be target data, so that the model can focus on the target information more.

We reduce the source data proportion with an exponential decay function. $\alpha$ represents the initial proportion of the source data in the training set. $t$ represents the current training step and $m$ represents the number of batches in total. $\lambda$ represents the exponential decay rate of $\alpha$. $r_S^t$ and $r_T^t$ represent the proportion of the source data and proportion of target data at time step $t$.

\begin{equation}
    r_S^t = \alpha  \lambda^{t-1}, 0 < \alpha < 1, 0 < \lambda < 1
\end{equation}
\begin{equation}
    r_T^t = 1 - \alpha \cdot \lambda^{t-1}
\end{equation}

Let $D_S$ represents the accumulated source data used to train the model, and let $B$ represents the batch size. We have 
\begin{equation}
    D_S = B \cdot \sum_{t=1}^{m}r_S^t = B \cdot \frac{\alpha\cdot(1-\lambda^m)}{1-\lambda}
\end{equation}
After the model is updated for adequate batches, we can approximate $D_S$ using
\begin{equation}
    D_S = B \cdot \frac{\alpha}{1-\lambda}
    \label{batch_size_formule}
\end{equation}

We could empirically estimate proper $D_S$ based on the relation between source dataset and target dataset. For example, the higher the similarity between the source and the target data, the larger the $D_S$ is. Because, the more similar the source data is to the target data, there is more knowledge the target task can borrow from the source task. If researchers want to simplify the hyperparameters tuning process or constrain the influence of source data, $\alpha$ can be set by $D_S$:
\begin{equation}
    \alpha = D_S \cdot (1 - \lambda) / B
    \label{equa:calculate DS}
\end{equation}


\section{Experimental Design}
We validate it by two popular model LSTM and BERT on three tasks: named entity recognition (NER), part-of-speech tagging (POS) and chunking. There tasks have much worse performance on formal text (such as news) than informal text (such as tweets). 

\subsection{Datasets}
\label{dataset_subsection}

We use OntoNotes-nw as the source dataset, and RitterNER dataset as the target dataset to validate NER task.
While we use Penn Treebank (PTB) POS tagging dataset as the source data set, and RitterPOS as the target dataset in the POS tagging task. For the chunking task, we use CoNLL 2000 as the source dataset, and RitterCHUNK as the target dataset.
Please refer to Appendix \ref{append:dataset statistic} for more details about datasets.


\subsection{Model Setting}

We implemented BERT and LSTM to validate the effect of data annealing on all three tasks. \par
\noindent\textbf{BERT}. We implemented both BERT\textsubscript{BASE} model and BERT\textsubscript{LARGE} model. We use CRF as a classifier on the top of BERT structure.
In some tasks, the source dataset and target dataset do not have the same set of labels. Therefore, we use two separate CRF classifiers for source task and target task.\par
\noindent\textbf{LSTM}. We used character and word embedding as input features as previous works
\citep{ncrf++, Yang-transfer}. 
We use one layer bidirectional LSTM to process the input features. as the same reason as in the implementation of BERT, we use two separate CRF classifiers on the top of LSTM structure. \par

We compare data annealing with two popular transfer learning paradigms, parameter initialization (INIT) and multi-task learning (MULT) \citep{transfer_learning_review,first_transfer_in_nlp}. Now we introduce the training strategy in experiments. \par

\noindent\textbf{Data annealing}. In all data annealing experiments, initial source data ratio $\alpha$ and decay rate $\lambda$ are tuned in range (0.9, 0.99). When training BERT model, we also calculated the estimated total batches from source data $D_S$ that fed into the model by equation \ref{equa:calculate DS}. By avoiding a large $D_S$, the model has lower probability to suffer from catastrophic forgetting as mentioned in section \ref{methods_section}. \par 

\noindent\textbf{MULT}. We implemented MULT on both LSTM-CRF and BERT-CRF structure. In all MULT experiments, following \citet{Yang-transfer} and \citet{multi_transfer_learning}, we tune the ratio of source data in range (0.1, 0.9).

\noindent\textbf{INIT}. We implemented INIT on BERT-CRF structure. In all INIT experiments, we run three times on source data and pick the model that achieves the highest performance. Then we continue to fine-tune the model on target task.

\begin{figure}
    \includegraphics[width=7.7cm]{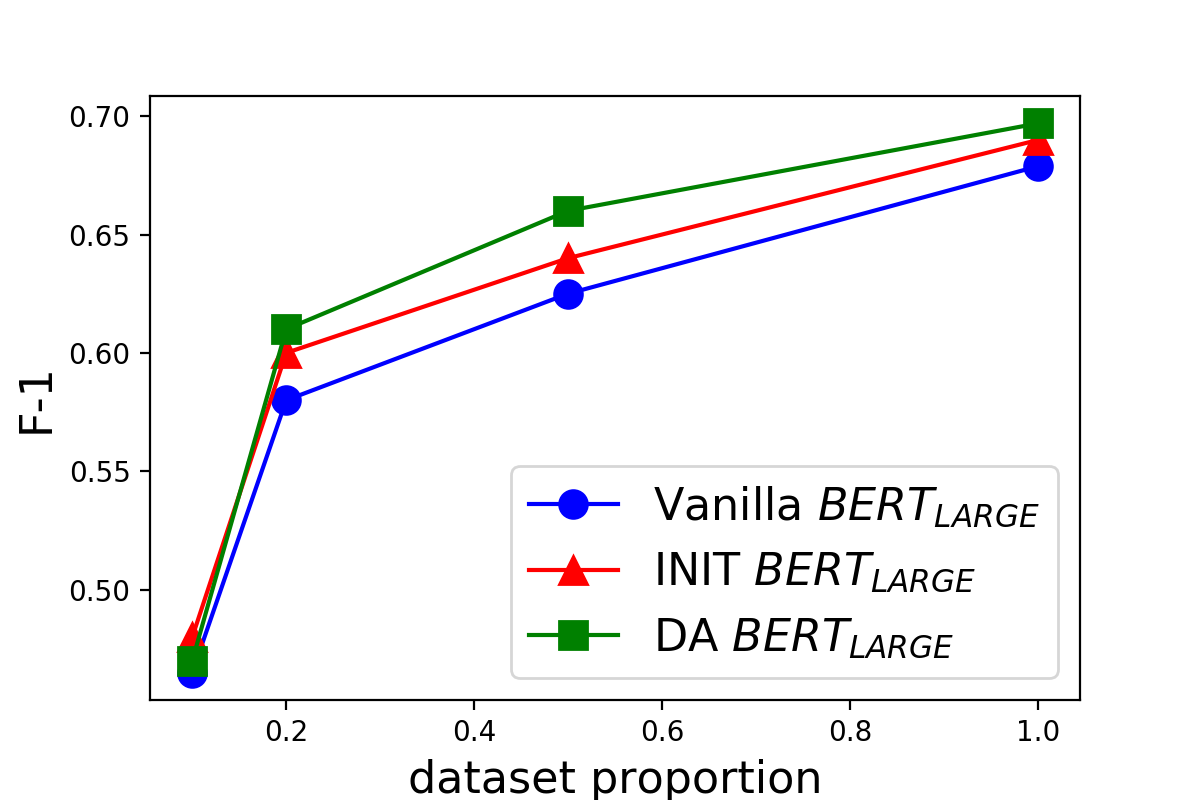}
    \caption{Performance on named entity recognition task. DA BERT\textsubscript{LARGE} indicates Vanilla BERT\textsubscript{LARGE} finetuned with data annealing.}
    \label{dataset_size_changing}
\end{figure}

\section{Experiment Results}
The result of three tasks are shown in Table \ref{tab:main_result}.
Vanilla means the model is trained without transfer learning, in other words, the model does not utilize the source data.
DA means the model is implemented with data annealing procedure.
All the numbers in the tables are the average result of three runs. It is worth noting that state-of-the-art results on these three tasks are achieved by different models. While our proposed data annealing algorithm is applied to the same simple BERT or LSTM structure without fancy decoration.

\begin{table*}[]
    \centering
    \begin{tabular}{ p{3.9cm}||p{1.20cm}|p{1.20cm}|p{1.20cm}|p{1.20cm}|p{1.20cm}|p{1.20cm}|p{1.20cm}  }
     \hline
     \multirow{2}{*}{model}  & \multicolumn{3}{c|}{NER} & POS & \multicolumn{3}{c}{Chunking} \\
     \cline{2-8}
     &$P$ & $R$ & $F_1$ & $A$ &$P$ & $R$ & $F_1$ \\
    \hline
     
     Vanilla LSTM & \textbf{75.55}  & 55.75 & 64.05& 88.65&  83.76 & 83.78  & 83.77\\
     MULT LSTM  & 74.51  & 58.48  & 65.49&88.81 &\textbf{83.92}  & 84.48  & 84.20   \\
     DA LSTM & 75.51 & \textbf{61.01} & \textbf{67.45} &\textbf{89.16}& 83.81 & \textbf{85.37}  & \textbf{84.58}\\
     \hline
     Vanilla BERT\textsubscript{BASE}  &  68.73 & 62.74 & 65.58 &91.05& 85.05 & 85.96 & 85.50 \\
     INIT BERT\textsubscript{BASE} &   69.28  &\textbf{63.74} &66.40  &90.85&  85.48  & 86.77  & 86.13  \\
     MULT BERT\textsubscript{BASE}  &  70.42   &62.38 & 66.12 &91.39 &  86.01  & 87.75  & 86.87  \\
     DA BERT\textsubscript{BASE} & \textbf{71.09} & \textbf{63.74}  & \textbf{67.21} &\textbf{91.55} &  \textbf{86.16} & \textbf{87.91} & \textbf{87.03}  \\
     \hline
     Vanilla BERT\textsubscript{LARGE}  & 68.41 &67.45 & 67.88 &91.88 & 85.55 & 86.78 & 86.16 \\
     INIT BERT\textsubscript{LARGE}  & 68.85 & \textbf{69.20} & 68.99  &92.04  & 86.42 & 87.59 & 87.00\\
     MULT BERT\textsubscript{LARGE} & 70.05  & 66.08  & 68.00 &92.06 & 86.29 & 87.21 & 86.54   \\
     DA BERT\textsubscript{LARGE}& \textbf{70.61} & 68.81   & \textbf{69.69} &\textbf{92.54} & \textbf{86.71} & \textbf{88.15}  & \textbf{87.53}   \\
     \hspace{3mm}*Over state-of-the-art & \textbf{\textcolor{red}{\hspace{1.5mm}-5.51}} & \textbf{\textcolor{green}{\hspace{1.5mm}+9.71}}   & \textbf{\textcolor{green}{\hspace{1.5mm}+3.16}} &\textbf{\textcolor{green}{\hspace{1.5mm}+1.37}} & \textbf{\textcolor{green}{\hspace{1.5mm}+2.24}} & \textbf{\textcolor{green}{\hspace{1.5mm}+3.61}}  & \textbf{\textcolor{green}{\hspace{1.5mm}+3.03}}   \\
     \hline
     **State-of-the-art  &76.12  &59.10   & 66.53 &91.17 &84.47 &84.54& 84.50 \\
     \hline
    \end{tabular}
    \caption{Results on NER, POS tagging and chunking task. * means the difference between DA BERT\textsubscript{LARGE} and state-of-the-art results. ** means the state-of-the-art for these three tasks are achieved by different models. \citet{Yang_ner_previous_best}, \citet{POS_previous_best} and \citet{Yang-transfer} proposed the state-of-the-art model on NER, POS tagging and chunking respectively.}
    \label{tab:main_result}
\end{table*}



\noindent\textbf{Named Entity Recognition (NER)}. Our annealing procedure outperforms other transfer learning procedures in terms of $F_1$, meaning our data annealing is especially effective in striking a balance between the precision and recall in extracting named entities in informal text.
Usually, a sentence contains more words that are not entities. So if the model is not sure whether a word is an entity, the model is likely to predict it as not an entity in order to reduce the training loss. The state-of-the-art models easily achieved the highest precision, but their recalls are lower. It indicates that the state-of-the-art methods achieve high performance by predicting fewer entities, while BERT models receive high performance by both covering more entities and predicting them correctly. 

\noindent\textbf{Part-of-speech Tagging (POS tagging).} LSTM achieves a higher accuracy under our proposed data annealing procedure compared to the other two transfer learning procedures. Both BERT\textsubscript{BASE} and BERT\textsubscript{LARGE} under our data annealing procedure outperform other transfer learning procedures.
The improvement over the state-of-the-art method is 1.37 in accuracy measure in POS tagging.

\noindent\textbf{Chunking.} When LSTM, BERT\textsubscript{BASE} and BERT\textsubscript{LARGE} are used as the training model under our data annealing procedure, they achieve better performances compared to other transfer learning paradigms.
Our best model outperforms the state-of-the-art model by 3.03 in $F_1$. 

\noindent\textbf{The Dataset Size Influence.}
To further evaluate our methods performance when there is limited labelled data, we randomly sample 10\%, 20\% and 50\% of the training set in RitterNER. Then we compare our proposed DA BERT\textsubscript{LARGE} with INIT BERT\textsubscript{LARGE} and Vanilla BERT\textsubscript{LARGE} baselines. The result in Figure \ref{dataset_size_changing} shows that our model is still better than INIT BERT\textsubscript{LARGE} on limited resources condition and achieves a significant improvement over Vanilla BERT\textsubscript{LARGE} baseline.

\section{Error Analysis}
We performed error analysis on the named entity task about RitterNER dataset.
We first calculated the $F_1$ score of the ten predefined entity types.
We find that compared with Vanilla BERT\textsubscript{LARGE} and INIT BERT\textsubscript{LARGE}, DA BERT\textsubscript{LARGE} achieves higher $F_1$ score on two frequent entities, {\small "PERSON"} and {\small "OTHER"}.  {\small "PERSON"} is a frequent concept in formal data. It shows our method learns to utilize formal data knowledge to obtain an improved {\small "PERSON"} detection.
Besides, {\small "OTHER"} means entities that are not in the ten predefined entity types. Higher performance on {\small "OTHER"} suggests DA BERT\textsubscript{LARGE} has a better understanding of the general concept of an entity.
INIT BERT\textsubscript{LARGE} achieves a higher $F_1$ score on another frequent entity type, {\small "GEO-LOC"}, showing the effectiveness of traditional transfer learning methods.
We did not find clear difference in other entities types.

Besides, we found that if a word belongs to a rarely appeared entity type, all the three models are less likely to predict this word's entity type correctly. We suspect that if a word belongs to a frequent seen entity type, then even if the exact work never appeared in the training set, another similar word that has a similar representation may be in the training data. So the model is able to predict a word by learning from other similar words that belong to the same type. We plan to assign more penalty to infrequent entity types to tackle this issue in the future.\par

We noticed that the improvement in recently reported literature on these tasks is usually less than 0.5 in absolute value in $F_1$. Considering the noisy nature of informal text data, we suspect the model is close to its theoretical maximum performance. To prove this, we randomly sampled 30 sentences 
We found that a fairly large proportion of sentences are too noisy to predict correctly. Please refer to Appendix~\ref{append:mispredicted example} for some examples.  This suggests that transfer learning has limited effect when the dataset has a strong noisy feature. A denoising technique could be useful in this scenario, and a pre-trained model based on noisy text could be another possible solution. 

\section{Conclusion}
In this paper, we propose data annealing, a model-independent transfer learning procedure for informal language understanding tasks. It is applicable to various models such as LSTM and BERT. It has been proven as a good approach to utilizing knowledge from formal data to informal data by exhaustive experiments. When data annealing is applied with BERT, it outperforms different state-of-the-art models on different informal language understanding tasks. Since large pre-trained models have been widely used, it could also serve as a good fine-tuning method. Data annealing is also effective when there is limited labelled resources.


\bibliography{acl2020}

\begin{thebibliography}{19}
\expandafter\ifx\csname natexlab\endcsname\relax\def\natexlab#1{#1}\fi

\bibitem[{Bengio et~al.(2009)Bengio, Louradour, Collobert, and
  Weston}]{curriculum_learning}
Yoshua Bengio, J{\'e}r\^{o}me Louradour, Ronan Collobert, and Jason Weston.
  2009.
\newblock \href {https://doi.org/10.1145/1553374.1553380} {Curriculum
  learning}.
\newblock In \emph{Proceedings of the 26th Annual International Conference on
  Machine Learning}, ICML '09, pages 41--48, New York, NY, USA. ACM.

\bibitem[{Bertsimas and Tsitsiklis(1993)}]{simulated_annealing}
Dimitris Bertsimas and John Tsitsiklis. 1993.
\newblock \href {https://doi.org/10.1214/ss/1177011077} {Simulated annealing}.
\newblock \emph{Statist. Sci.}, 8(1):10--15.

\bibitem[{Collobert and Weston(2008)}]{multi_transfer_learning}
Ronan Collobert and Jason Weston. 2008.
\newblock \href {https://doi.org/10.1145/1390156.1390177} {A unified
  architecture for natural language processing: Deep neural networks with
  multitask learning}.
\newblock In \emph{Proceedings of the 25th International Conference on Machine
  Learning}, ICML '08, pages 160--167, New York, NY, USA. ACM.

\bibitem[{Devlin et~al.(2018)Devlin, Chang, Lee, and Toutanova}]{bert-paper}
Jacob Devlin, Ming{-}Wei Chang, Kenton Lee, and Kristina Toutanova. 2018.
\newblock \href {http://arxiv.org/abs/1810.04805} {{BERT:} pre-training of deep
  bidirectional transformers for language understanding}.
\newblock \emph{CoRR}, abs/1810.04805.

\bibitem[{Gui et~al.(2018)Gui, Zhang, Gong, Peng, Liang, Ding, and
  Huang}]{POS_previous_best}
Tao Gui, Qi~Zhang, Jingjing Gong, Minlong Peng, Di~Liang, Keyu Ding, and
  Xuanjing Huang. 2018.
\newblock \href {https://doi.org/10.18653/v1/D18-1275} {Transferring from
  formal newswire domain with hypernet for twitter {POS} tagging}.
\newblock In \emph{Proceedings of the 2018 Conference on Empirical Methods in
  Natural Language Processing}, pages 2540--2549, Brussels, Belgium.
  Association for Computational Linguistics.

\bibitem[{Kshirsagar et~al.(2018)Kshirsagar, Cukuvac, McKeown, and
  McGregor}]{twitter_hate_speech}
Rohan Kshirsagar, Tyus Cukuvac, Kathleen~R. McKeown, and Susan McGregor. 2018.
\newblock \href {http://arxiv.org/abs/1809.10644} {Predictive embeddings for
  hate speech detection on twitter}.
\newblock \emph{CoRR}, abs/1809.10644.

\bibitem[{Lee et~al.(2018)Lee, Dernoncourt, and Szolovits}]{lstm_transfer_ner}
Ji~Young Lee, Franck Dernoncourt, and Peter Szolovits. 2018.
\newblock \href {https://www.aclweb.org/anthology/L18-1708} {Transfer learning
  for named-entity recognition with neural networks}.
\newblock In \emph{Proceedings of the Eleventh International Conference on
  Language Resources and Evaluation ({LREC}-2018)}, Miyazaki, Japan. European
  Languages Resources Association (ELRA).

\bibitem[{Mou et~al.(2016)Mou, Meng, Yan, Li, Xu, Zhang, and
  Jin}]{first_transfer_in_nlp}
Lili Mou, Zhao Meng, Rui Yan, Ge~Li, Yan Xu, Lu~Zhang, and Zhi Jin. 2016.
\newblock \href {http://arxiv.org/abs/1603.06111} {How transferable are neural
  networks in {NLP} applications?}
\newblock \emph{CoRR}, abs/1603.06111.

\bibitem[{Peters et~al.(2018)Peters, Neumann, Iyyer, Gardner, Clark, Lee, and
  Zettlemoyer}]{elmo}
Matthew~E. Peters, Mark Neumann, Mohit Iyyer, Matt Gardner, Christopher Clark,
  Kenton Lee, and Luke Zettlemoyer. 2018.
\newblock \href {http://arxiv.org/abs/1802.05365} {Deep contextualized word
  representations}.
\newblock \emph{CoRR}, abs/1802.05365.

\bibitem[{Peters et~al.(2019)Peters, Ruder, and Smith}]{pretrain_or_not}
Matthew~E. Peters, Sebastian Ruder, and Noah~A. Smith. 2019.
\newblock \href {http://arxiv.org/abs/1903.05987} {To tune or not to tune?
  adapting pretrained representations to diverse tasks}.
\newblock \emph{CoRR}, abs/1903.05987.

\bibitem[{Ruder et~al.(2017)Ruder, Ghaffari, and Breslin}]{data_selection}
Sebastian Ruder, Parsa Ghaffari, and John~G. Breslin. 2017.
\newblock \href {http://arxiv.org/abs/1702.02426} {Data selection strategies
  for multi-domain sentiment analysis}.
\newblock \emph{CoRR}, abs/1702.02426.

\bibitem[{Ruder and Plank(2017)}]{learning_to_select_data}
Sebastian Ruder and Barbara Plank. 2017.
\newblock \href {http://arxiv.org/abs/1707.05246} {Learning to select data for
  transfer learning with bayesian optimization}.
\newblock \emph{CoRR}, abs/1707.05246.

\bibitem[{Sun et~al.(2019)Sun, Qiu, Xu, and Huang}]{bert_forgetting}
Chi Sun, Xipeng Qiu, Yige Xu, and Xuanjing Huang. 2019.
\newblock \href {http://arxiv.org/abs/1905.05583} {How to fine-tune {BERT} for
  text classification?}
\newblock \emph{CoRR}, abs/1905.05583.

\bibitem[{van~der Wees et~al.(2017)van~der Wees, Bisazza, and
  Monz}]{data_select_translation}
Marlies van~der Wees, Arianna Bisazza, and Christof Monz. 2017.
\newblock \href {https://doi.org/10.18653/v1/D17-1147} {Dynamic data selection
  for neural machine translation}.
\newblock In \emph{Proceedings of the 2017 Conference on Empirical Methods in
  Natural Language Processing}, pages 1400--1410, Copenhagen, Denmark.
  Association for Computational Linguistics.

\bibitem[{Weiss et~al.(2016)Weiss, Khoshgoftaar, and
  Wang}]{transfer_learning_review}
Karl Weiss, Taghi~M. Khoshgoftaar, and DingDing Wang. 2016.
\newblock \href {https://doi.org/10.1186/s40537-016-0043-6} {A survey of
  transfer learning}.
\newblock \emph{Journal of Big Data}, 3(1):9.

\bibitem[{Yang and Zhang(2018)}]{ncrf++}
Jie Yang and Yue Zhang. 2018.
\newblock \href {http://arxiv.org/abs/1806.05626} {{NCRF++:} an open-source
  neural sequence labeling toolkit}.
\newblock \emph{CoRR}, abs/1806.05626.

\bibitem[{Yang et~al.(2019)Yang, Lu, and Zheng}]{Yang_ner_previous_best}
Wei Yang, Wei Lu, and Vincent~W. Zheng. 2019.
\newblock \href {http://arxiv.org/abs/1902.00184} {A simple
  regularization-based algorithm for learning cross-domain word embeddings}.
\newblock \emph{CoRR}, abs/1902.00184.

\bibitem[{Yang et~al.(2017)Yang, Salakhutdinov, and Cohen}]{Yang-transfer}
Zhilin Yang, Ruslan Salakhutdinov, and William~W. Cohen. 2017.
\newblock \href {http://arxiv.org/abs/1703.06345} {Transfer learning for
  sequence tagging with hierarchical recurrent networks}.
\newblock \emph{CoRR}, abs/1703.06345.

\bibitem[{Zeiler(2012)}]{adadelta_use_annealing_learning_rate}
Matthew~D. Zeiler. 2012.
\newblock \href {http://arxiv.org/abs/1212.5701} {{ADADELTA:} an adaptive
  learning rate method}.
\newblock \emph{CoRR}, abs/1212.5701.

\end{thebibliography}
\bibliographystyle{acl_natbib}

\newpage

\onecolumn
\appendix

\section{Mispredicted Sentences Examples on Named Entity Recognition Task}
\label{append:mispredicted example}

\begin{table*}[h]
    \centering
    \begin{tabular}{c|p{14.5cm}}
	\hline
    
    1&Is making me purchase windows\textbf{\{NO\_ENTITY, B-PRODUCT\}} , antivirus and office\textbf{\{NO\_ENTITY, B-PRODUCT\}} \\
    \hline
    2& ellwood\textbf{\{NO\_ENTITY, B-PERSON\}} 's sushi , a glass of pinot , \&quot; strokes\textbf{\{NO\_ENTITY, B-OTHER\}} of\textbf{\{NO\_ENTITY, I-OTHER\}} genius\textbf{\{NO\_ENTITY, I-OTHER\}} \&quot; by john wertheim\textbf{\{NO\_ENTITY, I-PERSON\}} , play at barksdale\textbf{\{NO\_ENTITY, B-FACILITY\}} in a bit , lovely friday night :) \\
    \hline
    3& lalala\textbf{\{B-GEO-LOC, NO\_ENTITY\}} south\textbf{\{B-GEO-LOC, NO\_ENTITY\}} game tonight !!!! Go us . http://bit.ly/b351o9
    RT \@BunBTrillOG : Okay \#teamtrill time to show them our power ! \#BunB106andPark needs to trend now ! RT til it hurts ! I got ya twitter\textbf{\{NO\_ENTITY, B-COMPANY\}} jail ... \\
    \hline
    4& Chicago Weekend Events : Lebowski\textbf{\{NO\_ENTITY, B-OTHER\}} Fest\textbf{\{NO\_ENTITY, I-OTHER\}} , Dave\textbf{\{NO\_ENTITY, B-PERSON\}} Matthews\textbf{\{NO\_ENTITY, I-PERSON\}} , Latin Music And More : The lively weekend ( well , Friday throu ... http://bit.ly/cLTnyl \\
    \hline
    5& RT @DonnieWahlberg : Soldiers ... Familia ... BH's...\textbf{\{B-PERSON, NO\_ENTITY\}} NK Fam ... Homies ... Etc . Etc . Etc .... I 'm gonna need some company next Friday in NYC ... \\
    \hline
    6& tell ur dad2bring the ypp back in Hayes\textbf{\{B-GEO-LOC, NO\_ENTITY\}} we sorted it out last time I'm like yea I'll tell him *covers eyes*wat informing am I doing \#llowit   \\
    \hline
    7& \#aberdeen RT \@flook\_firehose2010Polar Bear http://flook.it/c/1H1HZq Sun , 17 Oct 2010 at 10:28 am The Tunnels Carnegies\textbf{\{B-GEO-LOC, NO\_ENTITY\}} Brae Aberdeen\textbf{\{B-GEO-LOC, NO\_ENTITY\}} Un ... \\
    \hline
    8& \&lt; 3 it RT \@Djcheapshot : Tonite I m DJing at Mai\textbf{\{NO\_ENTITY, B-FACILITY\}} Tai\textbf{\{NO\_ENTITY, I-FACILITY\}} in Long Beach\textbf{\{B-GEO-LOC, I-GEO-LOC\}} . I'm considering wearing MY TIE !! Get it ? My tie = Mai Tai ? No ? Sorry . Bye . \\
    \hline
    9& \&quot; I gotta admit , Alex\textbf{\{NO\_ENTITY, B-PERSON\}} sounds hot when he talks in spanish during the ' Alejandro\textbf{\{NO\_ENTITY, B-OTHER\}} ' Cover \&quot; -via someone 's tumblr\textbf{\{NO\_ENTITY, B-COMPANY\}} I'm pleased to have introduced \@TheSmokingGunn to twitter\textbf{\{NO\_ENTITY, B-COMPANY\}} . May he become as inane as me . \\
    \hline
    10& Before I proceed into the paradise , let 's not forget the Princess\textbf{\{NO\_ENTITY, B-MOVIE\}} Lover\textbf{\{NO\_ENTITY, I-MOVIE\}} OVA\textbf{\{NO\_ENTITY, I-MOVIE\}} 1\textbf{\{NO\_ENTITY, I-MOVIE\}} teaser pic , SFW\textbf{\{B-GEO-LOC, NO\_ENTITY\}} http://yfrog.com/0fg2kfj \\
    \hline
    \end{tabular}
    \caption{Ten examples of mispredictted sentences. }
\end{table*}

\section{Dataset Statistic}
\label{append:dataset statistic}

\begin{table*}[!h]
    \centering
    \begin{tabular}{ p{1.9cm}||p{1.3cm}|p{2.0cm}|p{1.9cm}|p{1.9cm}|p{1.9cm}}
     \hline
     Task Type&  Category & Dataset& Train Tokens& Dev Tokens & Test Tokens \\
     \hline
     \multirow{2}{*}{NER} & Formal & Ontonote-nw &    848,220& 144,319&49,235   \\
     \cline{2-6}
     & Informal & RitterNER &37,098&4,461 &4,730   \\
     \hline
     \multirow{2}{*}{POS Tagging} & Formal & PTB 2003 & 912,344 &131,768 &129,654  \\
     \cline{2-6}
     & Informal & RitterPOS  &  10,857 &2,242 & 2,291 \\
     \hline
     \multirow{2}{*}{Chunking}   & Formal & CoNLL 2000 & 211,727& - & 47,377   \\
     \cline{2-6}
     & Informal & RitterCHUNK & 10,610 &2,309 &2,292  \\
     \hline
    \end{tabular}
    \caption{Dataset statistics.}
    \label{tab:dataset_stat}
\end{table*}

\clearpage
\end{document}